\documentclass{sig-alternate-05-2015}

\usepackage[breaklinks=true,bookmarks=false,colorlinks=true,backref]{hyperref}
\usepackage{flushend}

\makeatletter
\def\@copyrightspace{\relax}
\makeatother

\begin{document}

\title{CrowdNet: A Deep Convolutional Network for Dense Crowd Counting}

\numberofauthors{3} 

\author{
 \alignauthor Lokesh Boominathan \\
        \affaddr{Video Analytics Lab}\\
       \affaddr{Indian Institute of Science}\\
       \affaddr{Bangalore, INDIA - 560012}\\
       \affaddr{boominathanlokesh@gmail.com}
 \alignauthor Srinivas S S Kruthiventi \\
        \affaddr{Video Analytics Lab}\\
       \affaddr{Indian Institute of Science}\\
       \affaddr{Bangalore, INDIA - 560012}\\
       \affaddr{kssaisrinivas@gmail.com}
 \alignauthor R. Venkatesh Babu \\
        \affaddr{Video Analytics Lab}\\
       \affaddr{Indian Institute of Science}\\
       \affaddr{Bangalore, INDIA - 560012}\\
       \affaddr{venky@cds.iisc.ac.in}
 }

\maketitle
\begin{abstract}
 Our work proposes a novel deep learning framework for estimating crowd density from static images of highly dense crowds. We use a combination of deep and shallow, fully convolutional networks to predict the density map for a given crowd image. Such a combination is used for effectively capturing both the high-level semantic information (face/body detectors) and the low-level features (blob detectors), that are necessary for crowd counting under large scale variations. As most crowd datasets have limited training samples (<100 images) and deep learning based approaches require large amounts of training data, we perform multi-scale data augmentation. Augmenting the training samples in such a manner helps in guiding the CNN to learn scale invariant representations. Our method is tested on the challenging UCF\_CC\_50 dataset, and shown to outperform the state of the art methods.      
\end{abstract}


\section{Introduction}
In the light of problems caused due to poor crowd management, such as crowd crushes and blockages, there is an increasing need for computational models which can analyse highly dense crowds using video feeds from surveillance cameras. Crowd counting is a crucial component of such an automated crowd analysis system. This involves estimating the number of people in the crowd, as well as the distribution of the crowd density over the entire area of the gathering. Identifying regions with crowd density above the safety limit can help in issuing prior warnings and can prevent potential crowd crushes. Estimating the crowd count also helps in quantifying the significance of the event and better handling of logistics and infrastructure for the gathering.

In this work, we propose a deep learning based approach for estimating the crowd density as well as the crowd count from still images. Counting crowds in highly dense scenarios (>2000 people) poses a variety of challenges. Highly dense crowd images suffer from severe occlusion, making the traditional face/person detectors ineffective. Crowd images can be captured from a variety of angles introducing the problem of perspective. This results in non-uniform scaling of the crowd necessitating the estimation model to be scale-invariant to large scale changes. Furthermore, unlike other vision problems, annotating highly dense crowd images is a laborious task. This makes the creation of large-scale crowd counting datasets infeasible and limits the amount of training data available for learning-based approaches.

\begin{figure}[t]
    \centering
    \begin{minipage}{\linewidth}
        \centering
        \includegraphics[width=0.89\linewidth]{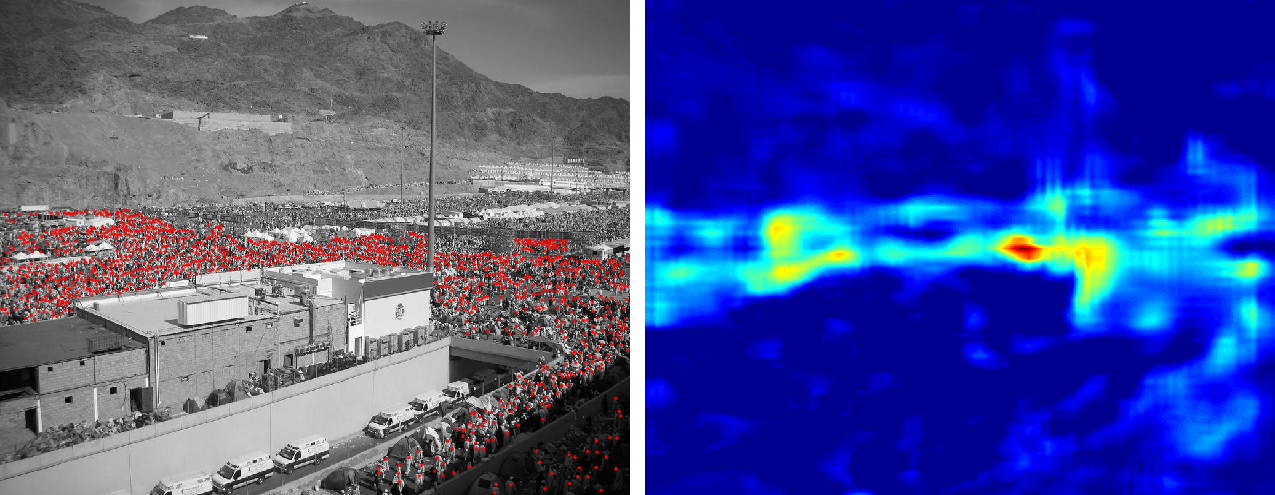}\\
        Actual Count: 1115 \hspace{1cm} Estimated: 1143
        \vspace{0.15cm}
    \end{minipage}%
    \\
    \begin{minipage}{\linewidth}
        \centering
        \includegraphics[width=0.89\linewidth]{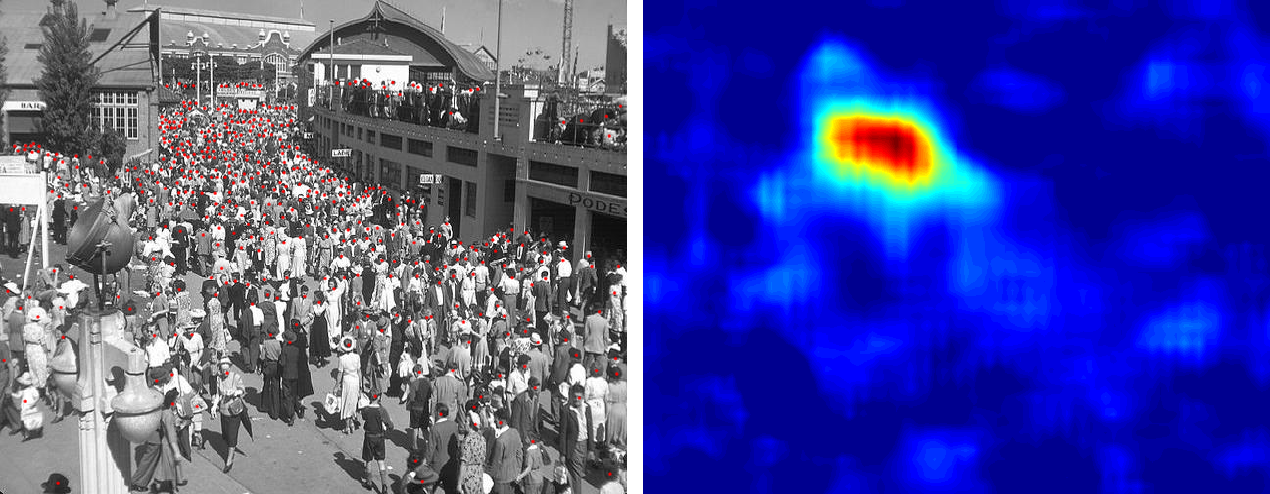}\\
        Actual Count:440 \hspace{1cm} Estimated:433
    \end{minipage}
    \caption{Crowd images with head annotations marked using red dots and their corresponding estimated crowd density maps}
    \label{fig:teaser}
\end{figure}

\begin{figure*}[ht]
    \includegraphics[width=\linewidth]{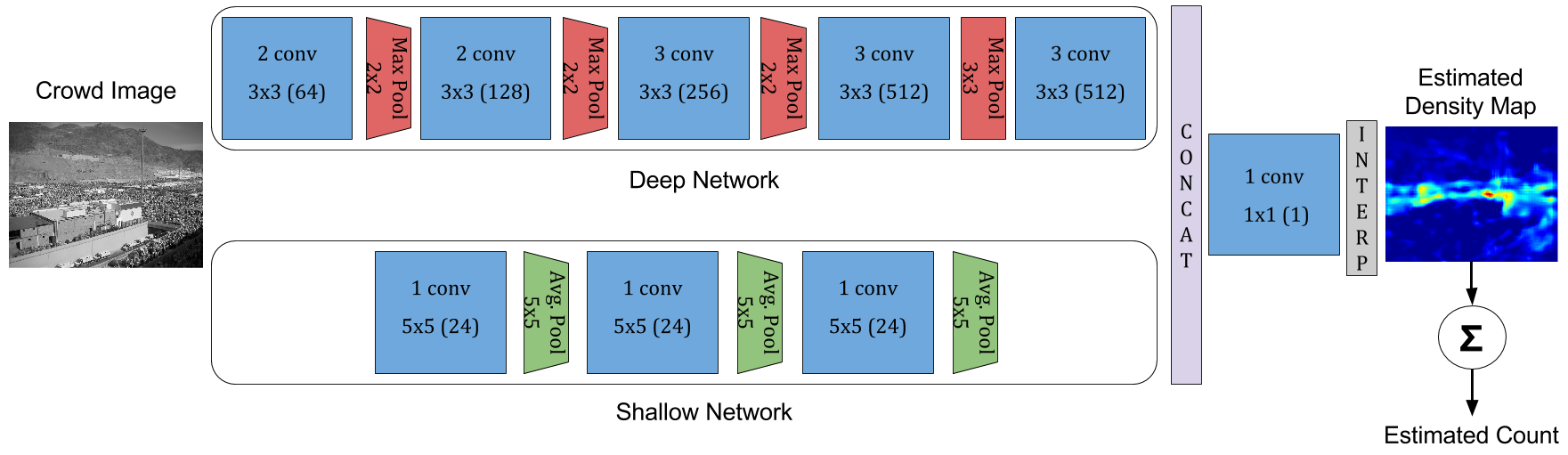}
    \caption{Overview of the proposed architecture for crowd counting}
    \label{fig:Architecture}
\end{figure*}

Hand-crafted image features (SIFT~\cite{lowe2004distinctive}, HOG etc.~\cite{dalal2005histograms}) often fail to provide robustness to challenges of occlusion and large scale variations. Our approach for crowd counting relies instead on deep learnt features using the framework of fully convolutional neural networks(CNN). 

We tackle the issue of scale variation in the crowd images using a combination of a shallow and deep convolutional architectures. Further, we perform extensive data augmentation by sampling patches from the multi-scale image representation to make the system robust to scale variations. Our approach is evaluated on the challenging UCF\_CC\_50 dataset~\cite{UCFCC_dataset} and has achieved state of the art results.

\section{Related Work}

Some works in the crowd counting literature experiment on datasets having sparse crowd scenes~\cite{UCSD_data, kong2006viewpoint}, such as UCSD dataset~\cite{UCSD_data}, Mall dataset~\cite{MALL_dataset} and  PETS dataset~\cite{PETS_dataset}. In contrast, our method has been evaluated on highly dense crowd images which pose the challenges discussed in the previous section. Methods introduced in \cite{crowd_motion_1} and \cite{crowd_motion_2} exploit patterns of motion to estimate the count of moving objects. However, these methods rely on motion information which can be obtained only in the case of continuous video streams with a good frame rate, and do not extend to still image crowd counting. 

The algorithm proposed by Idrees \textit{et al.}~\cite{UCFCC_dataset} is based on the understanding that it is difficult to obtain an accurate crowd count using a single feature. To overcome this, they use a combination of handcrafted features: HOG based head detections, Fourier analysis, and interest points based counting. The post processing is done using multi-scale Markov Random Field. However, handcrafted features often suffer a drop in accuracy when subjected to variances in illumination, perspective distortion, severe occlusion etc. 

Though Zhang \textit{et al.}~\cite{Cross_scene} utilize a deep network to estimate crowd count, their model is trained using perspective maps of images. Generating these perspective maps is a laborious process and is infeasible. We use a simpler approach for training our model, yet obtain a better performance. Wang \textit{et al.}~\cite{new_ACMM} also train a deep model for crowd count estimation. Their model however is trained to determine only the crowd count and not the crowd density map, which is crucial for crowd analysis. Our network estimates both the crowd count as well as the crowd density distribution.

\section{Proposed Method}

\subsection{Network Architecture}

Crowd images are often captured from varying view points, resulting in a wide variety of perspectives and scale variations. People near the camera are often captured in a great level of detail i.e., their faces and at times their entire body is captured. However, in the case of people away from camera or when images are captured from an aerial viewpoint, each person is represented only as a head blob. Efficient detection of people in both these scenarios requires the model to simultaneously operate at a highly semantic level (faces/body detectors) while also recognizing the low-level head blob patterns. Our model achieves this using a combination of deep and shallow convolutional neural networks. An overview of the proposed architecture is shown in Fig.~\ref{fig:Architecture}. In the following subsections, we describe these networks in detail.

\subsubsection{Deep Network}

Our deep network captures the desired high-level semantics required for crowd counting using an architectural design similar to the well-known VGG-16~\cite{VGG} network. Although the VGG-16 architecture was originally trained for the purpose of object classification, the learned filters are very good generic visual descriptors and have found applications in a wide variety of vision tasks such as saliency prediction~\cite{kruthiventi2015deepfix}, object segmentation~\cite{deeplab1} etc. Our model efficiently builds up on the representative power of the VGG network by fine-tuning its filters for the problem of crowd counting. However, crowd density estimation requires per-pixel predictions  unlike the problem of image classification, where a single discrete label is assigned for an entire image. We obtain these pixel-level predictions by removing the fully connected layers present in the VGG architecture, thereby making our network fully convolutional in nature. 

The VGG network has 5 max-pool layers each with a stride of 2 and hence the resultant output features have a spatial resolution of only $1/32$ times the input image. In our adaptation of the VGG model, we set the stride of the fourth max-pool layer to $1$ and remove the fifth pooling layer altogether. This enables the network to make predictions at $1/8$ times the input resolution. We handle the receptive-field mismatch caused by the removal of stride in the fourth max-pool layer using the technique of holes introduced in \cite{deeplab}. Convolutional filters with holes can have arbitrarily large receptive fields irrespective of their kernel size. Using holes, we double the receptive field of convolutional layers after the fourth max-pool layer, thereby enabling them to operate with their originally trained receptive field.

\subsubsection{Shallow Network}

In our model, we aim to recognize the low-level head blob patterns, arising from people away from the camera, using a shallow convolutional network. Since blob detection does not require the capture of high-level semantics, we design this network to be shallow with a depth of only 3 convolutional layers. Each of these layers has $24$ filters with a kernel size of $5\times5$. To make the spatial resolution of this network's prediction equal to that of its deep counterpart, we use pooling layers after each convolution layer. Our shallow network is primarily used for the detection of small head-blobs. To ensure that there is no loss of count due to max-pooling, we use average pooling layers in the shallow network.

\subsubsection{Combination of Deep and Shallow Networks}

We concatenate the predictions from the deep and shallow networks, each having a spatial resolution of $1/8$ times the input image, and process it using a 1x1 convolution layer. The output from this layer is upsampled to the size of the input image using bilinear interpolation to obtain the final crowd density prediction. The total count of the people in the image can be obtained by a summation over the predicted density map. The network is trained by back-propagating the $l$2 loss computed with respect to ground-truth.

\subsection{Ground Truth}

Training a fully convolutional network using the ground-truth of head annotations, marked as a binary dot corresponding to each person, would be difficult. The exact position of the head annotations is often ambiguous, and varies from annotator to annotator (forehead, centre of the face etc.), making CNN training difficult. 

In \cite{new_ACMM}, the authors have trained a deep network to predict the total crowd count in an image patch. But using such a ground truth would be suboptimal, as it wouldn't help in determining which regions of the image actually contribute to the count and by what amount. Zhang \textit{et al.}~\cite{Cross_scene} have generated ground truth by blurring the binary head annotations, using a kernel that varies with respect to the perspective map of the image. However, generating such perspective maps is a laborious task and involves manually labelling several pedestrians by marking their height.

We generate our ground truth by simply blurring each head annotation using a Gaussian kernel normalized to sum to one. This kind of blurring causes the sum of the density map to be the same as the total number of people in the crowd. Preparing the ground truth in such a fashion makes the ground truth easier for the CNN to learn, as the CNN no longer needs to get the exact point of head annotation right. It also provides information on which regions contribute to the count, and by how much. This helps in training the CNN to predict both the crowd density as well as the crowd count correctly. 

\subsection{Data Augmentation}

As CNNs require a large amount of training data, we perform an extensive augmentation of our training dataset. We primarily perform two types of augmentation. The first type of augmentation helps in tackling the problem of scale variations in crowd images, while the second type improves the CNN's performance in regions where it is highly susceptible to making mistakes i.e., highly dense crowd regions. 

\begin{figure}[!h]
    \centering
    \includegraphics[width=0.92\linewidth]{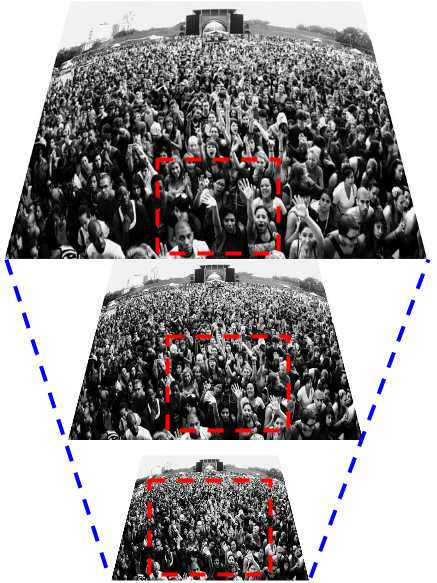}
    \caption{Our network is designed to be robust to scale variations by training it with patches cropped from multi-scale image pyramid.}
    \label{fig:ImagePyramid_1}
\end{figure}

In order to make the CNN robust to scale variations, we crop patches from the multi-scale pyramidal representation of each training image. We consider scales of $0.5$ to $1.2$, incremented in steps of $.1$, times the original image resolution (as shown in Fig.\ref{fig:ImagePyramid_1}) for constructing the image pyramid. We crop $225\times225$ patches with $50\%$ overlap from this pyramidal representation. With this augmentation, the CNN is trained to recognize people irrespective of their scales.

We observed that CNNs find highly dense crowds inherently difficult to handle. To overcome this, we augment the training data by sampling high density patches more often.

\section{Experiments}

We evaluate our approach for crowd counting on the challenging UCF\_CC\_50~\cite{UCFCC_dataset} dataset. This dataset contains 50 gray scale images, each provided with head annotations. The number of people per image varies between 94 and 4543, with an average of 1280 individuals per image. The dataset comprises of images from a wide range of scenarios such as concerts, political rallies, religious gatherings, stadiums etc.

In a manner similar to recent works \cite{Cross_scene, UCFCC_dataset}, we evaluate the performance of our approach using 5-fold cross validation. We randomly divide the dataset into five splits with each split containing 10 images. In each fold of the cross validation, we consider four splits (40 images) for training the network and the remaining split (10 images) for validating its performance. We sample $225\times225$ patches from each of the 40 training images following the previously described data augmentation method. This procedure yields an average of 50,292 training patches per fold. We train our deep convolutional network using the Deeplab~\cite{deeplab1, deeplab2} version of Caffe~\cite{caffe} deep learning framework, using Titan X GPUs. Our network was trained using Stochastic Gradient Descent (SGD) optimization with a learning rate of $1e-7$  and momentum of $0.9$. The average training time per fold is about 5 hours.

\subsection{Results}

We use Mean Absolute Error (MAE) to quantify the performance of our method. MAE computes the mean of absolute difference between the actual count and the predicted count for all the images in the dataset. The results of the proposed approach along with other recent methods are shown in Table.~\ref{tab:results}. The results shown do not include any post-processing methods. The results illustrate that our approach achieves state-of-the-art performance in crowd counting. 

\begin{table}[!htbp]
\centering
\label{tab:results}
\begin{tabular}{|c||c|}
\hline
Method & Mean Absolute Error \\
\hline\hline
Learning to Count~\cite{lempitsky2010learning} &	493.4 \\\hline
Density-aware Detection~\cite{rodriguez2011density} & 655.7 \\\hline
FHSc~\cite{UCFCC_dataset}	& 468.0 \\\hline
Cross-Scene Counting~\cite{Cross_scene} & 467.0 \\\hline
\textbf{Proposed} & \textbf{ 452.5}\\
\hline
\end{tabular}
\caption{Quantitative results of our approach along with other state-of-the-art methods on UCF\_CC\_50 Dataset.}
\end{table}

\begin{figure}[!h]
    \centering
    \includegraphics[width=\linewidth]{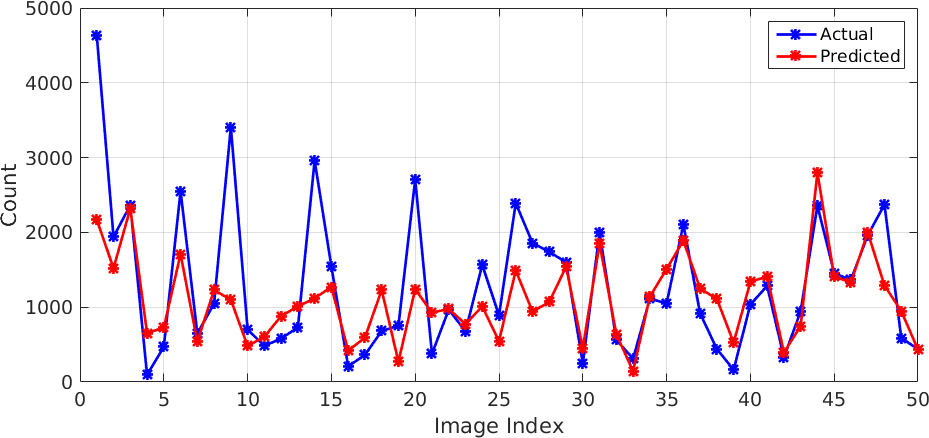}
    \caption{Actual count vs. Predicted Count for each of the 50 images in the UCF\_CC\_50 dataset.}
    \label{fig:pred_vs_act}
\end{figure}

We also show the predicted count for each image in the dataset along with its actual count in Fig.~\ref{fig:pred_vs_act}. For most of the images, the predicted count lies close to the actual count. However, we observe that the proposed approach tends to underestimate the count in cases of images with more than 2500 people. This estimation error could possibly be a consequence of the insufficient number of training images with such large crowds in the dataset.

\subsection{Analysis}

In this section, we analyse the following aspects of our approach using the hardest of the 5-folds.

\textbf{Deep and Shallow Networks:} Here, we experimentally show that combining both the deep and shallow networks, effectively captures individuals at multiple scales, thereby reducing the MAE. The experiment was performed on the hardest of 5 folds, and it was observed that using a combination of shallow and deep network gives a quantitative improvement over using just either one of them, as shown in Table~\ref{tab:comb_net}.

\begin{table}[!htbp]
\centering
\label{tab:results2}
\begin{tabular}{|c||c|}
\hline
Method & Mean Absolute Error \\
\hline\hline
Shallow Network & 1107	 \\\hline
Deep Network & 681\\\hline
Proposed (Deep + Shallow) & 645\\\hline
\end{tabular}
\caption{Quantitative results on the performance of the individual deep and shallow networks for crowd counting as opposed to the combined network, evaluated on the hardest of 5 folds.}
\label{tab:comb_net}
\end{table}

\textbf{Count based Augmentation:} Augmenting the training samples in favour of highly dense patches is observed to be effective at mitigating the lack of sufficient training samples with large crowds. Augmenting in such a fashion for the hardest fold, almost doubles the number of patches from 26,385 to 50,891. The quantitative advantage obtained by this augmentation is shown in the Table~\ref{tab:count_augmentation}. 

\begin{table}[!htbp]
\centering
\label{tab:results3}
\begin{tabular}{|c||c|}
\hline
Method & Mean Absolute Error \\
\hline\hline
Without augmentation & 725 \\\hline
Proposed (with augmentation) & 645 \\\hline
\end{tabular}
\caption{Quantitative results showing the advantage of augmenting data in favour of highly dense crowd patches, evaluated on the hardest of 5 folds.}
\label{tab:count_augmentation}
\end{table}

\section{Conclusion}
In this paper, we proposed a deep learning based approach to estimate the crowd density and total crowd count from highly dense crowd images. We showed that using a combination of a deep network as well as a shallow network is essential for detecting people under large scale variations and severe occlusion. We also show that the challenge of varying scales, and inherent difficulties in highly dense crowds, can be effectively tackled by augmenting the training images. Our method outperforms the state-of-the-art methods on the challenging UCF\_CC\_50 dataset.

\section{Acknowledgments}
This work was supported by Science and Engineering Research Board (SERB), Department of Science and Technology (DST), Govt. of India (Proj No. SB/S3/EECE/0127/2015).
\bibliographystyle{abbrv}

\end{document}